\documentclass[sigconf]{acmart}

\AtBeginDocument{%
  \providecommand\BibTeX{{%
    \normalfont B\kern-0.5em{\scshape i\kern-0.25em b}\kern-0.8em\TeX}}}

\usepackage{times}
\usepackage{url}
\usepackage{epsfig}
\usepackage{graphicx}
\usepackage{amsmath}
\usepackage{amssymb}
\usepackage{color}
\usepackage{todonotes}
\usepackage{multirow}
\usepackage{latexsym}
\usepackage{xcolor}
\usepackage{booktabs}
\usepackage{soul}
\usepackage{adjustbox}
\usepackage{tikz}
\usepackage{xspace}
\usepackage{wrapfig,lipsum,booktabs}

\copyrightyear{2020}
\acmYear{2020}
\setcopyright{acmcopyright}\acmConference[MM '20]{Proceedings of the 28th ACM International Conference on Multimedia}{October 12--16, 2020}{Seattle, WA, USA}
\acmBooktitle{Proceedings of the 28th ACM International Conference on Multimedia (MM '20), October 12--16, 2020, Seattle, WA, USA}
\acmPrice{15.00}
\acmDOI{10.1145/3394171.3414053}
\acmISBN{978-1-4503-7988-5/20/10}

\usepackage{enumitem}
\usepackage{gensymb}
\usepackage{booktabs}
\usepackage{subfigure}
\usepackage{multirow}
\usepackage{color}
\usepackage{mathtools}
\usepackage{amsthm}
\usepackage{soul}
\usepackage{algorithm}
\usepackage{algorithmic}
\usepackage{url}
\usepackage{bbm}

\newcommand{\R}{\mathbb{R}}

\acmSubmissionID{mmfp2297}
\begin{document}
\fancyhead{}

\title{Sequential Attention GAN for Interactive Image Editing}

\author{Yu Cheng$^1$,\quad Zhe Gan$^1$,\quad Yitong Li$^2$,\quad Jingjing Liu$^1$,\quad Jianfeng Gao$^3$}
\affiliation{$^1$Microsoft Dynamics 365 AI Research \quad $^2$Duke University \quad $^3$Microsoft Research}
\email{{yu.cheng,zhe.gan,jingjl,jfgao}@microsoft.com,yitong.li@duke.edu}

\newcommand{\JJ}[1]{\textcolor{blue}{\bf\small [JJ:#1]}}

\begin{abstract}
Most existing text-to-image synthesis tasks are static single-turn generation, based on pre-defined textual descriptions of images. To explore more practical and interactive real-life applications, we introduce a new task - Interactive Image Editing, where users can guide an agent to edit images via multi-turn textual commands on-the-fly. In each session, the agent takes a natural language description from the user as the input, and modifies the image generated in previous turn to a new design, following the user description. The main challenges in this sequential and interactive image generation task are two-fold: 1) contextual consistency between a generated image and the provided textual description; 2) step-by-step region-level modification to maintain visual consistency across the generated image sequence in each session. To address these challenges, we propose a novel Sequential Attention Generative Adversarial Network (SeqAttnGAN), which applies a neural state tracker to encode the previous image and the textual description in each turn of the sequence, and uses a GAN framework to generate a modified version of the image that is consistent with the preceding images and coherent with the description. To achieve better region-specific refinement, we also introduce a sequential attention mechanism into the model. To benchmark on the new task, we introduce two new datasets, Zap-Seq and DeepFashion-Seq, which contain multi-turn sessions with image-description sequences in the fashion domain. Experiments on both datasets show that the proposed SeqAttnGAN model outperforms state-of-the-art approaches on the interactive image editing task across all evaluation metrics including visual quality, image sequence coherence and text-image consistency.
\end{abstract}

\ccsdesc[500]{Information systems~Multimedia and multimodal retrieval}

\keywords{Generative Adversarial Network, Sequential Attention, Image Editing with Natural Language}

\maketitle

\begin{figure}[t!]
\centering
\includegraphics[width=0.36\textwidth]{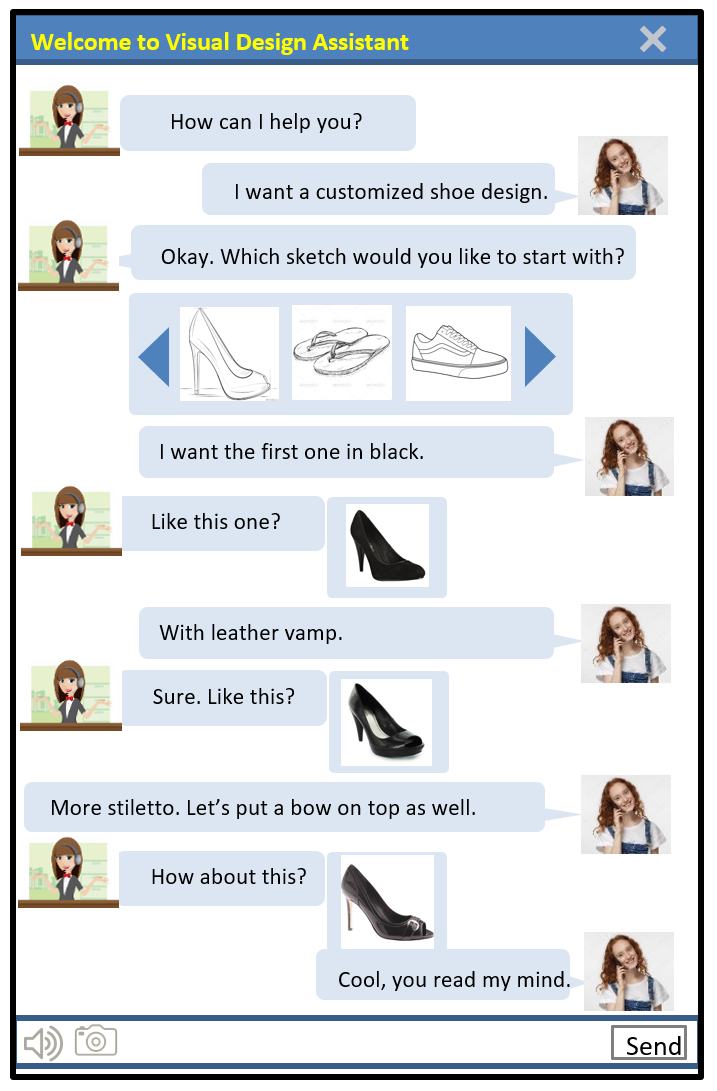}
\vspace{-1mm}
\caption{\small{Example of a visual design assistant powered by interactive image editing. In each sequence turn, the user provides natural language feedback to guide the system to modify the design. The system refines the images iteratively based on the user's feedback.}}
\label{fig:demo_system}
\vspace{-1mm}
\end{figure} 

\section{Introduction}
In recent years we have witnessed a tremendous growth in visual media, which has intensified users' needs for professional image editing tools (e.g., Adobe Photoshop, Microsoft Photos). However, image/video editing relies heavily on manual effort and is time-consuming, as visual design requires not only expert artistic creativity but also iterative experimentation through trial and error. To automate the process and save human effort, an AI-powered interactive design environment would allow a system to automatically generate new designs by following users' command through a multimodal interactive session. 

To reach the ultimate goal of enabling this creative collaboration between designers and algorithms, we propose a new task to approximate the setting and benchmark models - interactive image editing, where a system can generate new images by engaging with users in an interactive sequential setting. Figure \ref{fig:demo_system} illustrates an interactive image editing system, which supports natural language communication with a user for customizing shape, color, size, texture of a visual design through conversations. Users can provide feedback on intermediate results, which in turn allows the system to further refine the images.
Potential applications of such a system can go beyond visual design and extend to language-guided visual assistance/navigation. 

There are some related studies that explored similar tasks. For example,~\cite{Benmalek2018TheNP,Dixit_Kwitt_Niethammer_Vasconcelos_2017,libe,fashiongan,tagan} proposed approaches that allow systems to take keyword input (e.g., object attributes) for image generation. While these paradigms are effective to some degree, they are either restricted to keyword input or single-turn setting. Allowing only keywords inevitably constrains how much information a user can convey to the system to influence the image generation process. Furthermore, without multi-turn capability, the degree of interactive user experience with system assistance is very limited.

To solve these challenges, we propose a new conditional Generative Adversarial Network (GAN) framework, which uses an image generator to modify images following textual descriptions, and a neural state tracker to ensure the consistency of sequential context. In each turn, the generator generates a new image by taking into account both the history of previous textual descriptions and previously generated images. To fully preserve the sequential information in the image editing process, the model is trained end-to-end with full sequence sessions. To achieve better fine-grained image quality and coherent region-specific refinement, the model also uses an attention mechanism and a multimodal regularizer based on image-text matching score. 

As this is a newly proposed task, we introduce two new datasets, Zap-Seq and DeepFashion-Seq, which were collected via crowdsourcing in a real-world application scenario. In total, there are 8,734 collected sessions in Zap-Seq and 4,820 in DeepFashion-Seq. Each session consists of a sequences of images, with slight variation in design, accompanied by a sequence of sentences describing the difference between each pair of consecutive images. Figure \ref{fig:demo_system} shows an application powered by interactive image editing.

Experiments on these two datasets show that the proposed SeqAttnGAN framework achieves better performance than state-of-the-art techniques. In particular, by incorporating context history, SeqAttnGAN is able to generate high-quality images, beating all baseline models on metrics over contextual relevance and consistency. Detailed qualitative analysis and user study also show that allowing natural language feedback in image editing task is more effective than taking only keywords or visual attributes as input, which was used in previous approaches. The contributions of our work can be summarized as follows:
\begin{itemize}
\item We propose a new task - interactive image editing, which allows an agent to interact with a user for iterative image editing via multi-turn sequential interactions.
\item We introduce two new datasets for this task, Zap-Seq and DeepFashion-Seq. Consisting of image sequences paired with free-formed descriptions in diverse vocabularies, the two sets provide new benchmarks for measuring sequential image editing models. 
\item We propose a new conditional GAN framework, SeqAttnGAN, which can fully utilize context history to synthesize images that conform to users' iterative feedback in a sequential fashion. 
\end{itemize}

\section{Related Work}
\subsection{Image Generation and Editing}
Language-based image editing \cite{libe,manuvinakurike_conversational_2018} is a task designed for minimizing labor work while helping users create visual data. One big challenge is that systems should be able to understand which part of the image the user is referring to given an editing command. To achieve this, the system is required to have a comprehensive understanding of both natural language information and visual clues.
For example, Hu \textit{et al.} \cite{hu2016segmentation} focused on language-based image segmentation task, taking phrase as the input. Manuvinakurike \textit{et al.} \cite{manuvinakurike_conversational_2018} developed a system using simple language to modify the image, where a classification model is used to understand the user intent. 

Since the introduction of GAN \cite{gan,mmdgan}, there has been growing interest in image generation. In the conditional GAN space, there are studies on generating images from source image \cite{imagetrans,Liu2017UIT,pathakCVPR16context,CycleGAN2017}, sketch~\cite{sangkloy2016scribbler,NIPS2017_6650,xian2017texturegan}, scene graph~\cite{sg2im,ashual2019specifying}, object layout~\cite{zhao2019image,li2020bachgan}, or text (e.g., captions \cite{pmlr-v48-reed16}, attributes \cite{Dixit_Kwitt_Niethammer_Vasconcelos_2017}, long-paragraph \cite{li2018storygan}). There is also exploration on how to parameterize the model and training framework \cite{Mirza2014ConditionalGA} beyond the vanilla GAN \cite{pmlr-v70-odena17a}. Zhang \textit{et al.} \cite{stackgan} stacked several GANs for text-to-image synthesis, with different GANs generating images of different sizes. 

AttnGAN \cite{attngan} introduced attention mechanism into the generator, to focus on fine-grained word-level information. Chen \textit{et al.} \cite{libe} presented a framework targeting image segmentation and colorization with a recurrent attentive model. FashionGAN \cite{fashiongan} aimed at creating new clothing over a human body based on textual descriptions. The text-adaptive GAN~\cite{tagan} proposed a method for manipulating images with natural language description. While these paradigms are effective, they all have certain restrictions on the user input (either pre-defined attributes or single-turn interaction), which limits the scope of image editing applications.  
\subsection{Sequential Vision Tasks}
There are many vision+language tasks that lie in the intersection between computer vision and natural language processing, such as visual question-answering \cite{VQA}, visual-semantic embeddings \cite{wang2016learning}, grounding phrases in image regions \cite{rohrbach16eccv}, and image-grounded conversation \cite{mostafazadeh2017image}. 

Most approaches have focused on end-to-end neural models based on the encoder-decoder architecture and sequence-to-sequence learning \cite{gaosurvey,Serban:2016:BED:3016387.3016435,Bordes2016LearningEG}. Specifically, Das \textit{et al.} \cite{das2017visual} proposed the visual dialog task, where the agent aims to answer questions about an image in an interactive dialog. Vries \textit{et al.} \cite{DBLP:conf/cvpr/VriesSCPLC17} introduced the GuessWhat?! game, where a series of questions are asked to pinpoint a specific object in an image. However, these dialogue settings are mainly text-based, where visual features only play a complementary role. Manuvinakurike \textit{et al.} \cite{manuvinakurike_using_2017} investigated building dialog systems that can help users efficiently explore data through visualization. Guo \textit{et al.} \cite{DBLP:journals/corr/abs-1805-00145} introduced an agent that presents candidate images to the user and retrieves new images based on user's feedback. Another piece of related work is Benmalek \textit{et al.} \cite{Benmalek2018TheNP} on interactive image generation by encoding dialog history information. Different from these studies, in our work, text information is heavily relied on for guiding the image editing process throughout each image editing session. 

Compared to recent work on continuous image editing such as ChatPainter \cite{chatpainter}, our new datasets are designed for multi-turn image generation instead of single-turn. In addition, our data are derived from real fashion images, while CoDraw \cite{codraw} is based on cartoon images (Abstract Scenes dataset).

\section{Benchmarks: Zap-Seq and DeepFashion-Seq}
The interactive image editing task is defined as follows: in the $t$-th turn, the system presents a generated image $\hat{x}_t$ to the user, who then provides a textual feedback $o_t$ to describe the change he/she likes to make to realize a target design. The system then takes into account the user's feedback and generates a new image by modifying the previously generated image from the last turn. 
This process carries on iteratively until the user is satisfied with the result rendered by the system, or the maximum number of editing turns has been reached. 

Existing image generation datasets are mostly single-turned, thus not suitable for this sequential editing task.
To provide reliable benchmarks for the new task, we introduce two new datasets - Zap-Seq and DeepFashion-Seq, collected through crowdsourcing via Amazon Mechanical Turk (AMT)~\cite{amt}.

\begin{table}
\begin{center}
\small
\begin{tabular}{|c|c|c|}
\hline
Dataset & Zap-Seq & DeepFashion-Seq \\ 
\hline \hline
\# session & 8,734 & 4,820 \\
\# turns per session & 3.41 & 3.25 \\
\# descriptions & 18,497 & 12,765 \\
\# words per description &  6.83 & 5.79 \\
\# unique words & 973 & 687 \\
\hline
\end{tabular}
\end{center}
\caption{\small{Statistics on the Zap-Seq and DeepFashion-Seq datasets.}}\label{table:stat:dataset}
\end{table}

The optimal way to construct a sequential dataset for this task is collecting continuous dialog turns via a real interactive system. However, continuous dialog turns are difficult to collect, as the inherent nature of crowdsourcing limits the quality of complex data collection tasks. Real-time interactive interface on crowdsourcing platforms such as AMT is also difficult to control among a large pool of annotators. To circumvent the cumbersome and costly process of collecting pseudo human-machine interactions, we leverage two existing datasets - UT-Zap50K \cite{Yu:2014:FVC:2679600.2680101} and DeepFashion \cite{liuLQWTcvpr16DeepFashion}. UT-Zap50K contains 50,025 shoe images collected from
Zappos.com, and DeepFashion contains around 290,000 clothes images from different settings (e.g., store layouts, street snapshots). Each image is accompanied with a list of reference attributes. We first retrieve sequences of images from Zappos and DeepFashion datasets, with each sequence containing 3 to 5 image and every pair of consecutive images slightly different in certain attributes, as candidate sequences from interactive image editing sessions \cite{Zhao2017MemoryAugmentedAM}. The attribute manipulation procedure focuses on significant changes (colors, styles) first and finer-grained features (patterns, accessories) later, which is close to human behavior in realistic iterative image editing. As a result, a total of 8,734 image sequences were extracted from UT-Zap50K and 4,820 sequences from DeepFashion as the image sequence pool. 

After collecting the image sequences, the second step is to pair them with natural language descriptions that can capture the differences between each image pair, in order to mimic sequential interactive editing sessions. For this, we resort to crowdsourcing via AMT. Specifically, each annotator was asked to provide a free-formed sentence to describe the differences between any two given images. To promote specific and relevant textual descriptions and mimic the real natural language interactive environment, we also provide sentence prefixes for the annotators to select and complete when composing their responses to a retrieved image.

Figure \ref{fig:data_collection} provides some image sequence examples with textual annotations collected from the turkers (more examples are provided in Appendix). To provide robust datasets for benchmarking, we also randomly select a subset of images from the two original datasets as additional sequences 
to be annotated in AMT, which make up to $10\%$ of the whole datasets.

\begin{figure}
\centering
\includegraphics[width=0.48\textwidth]{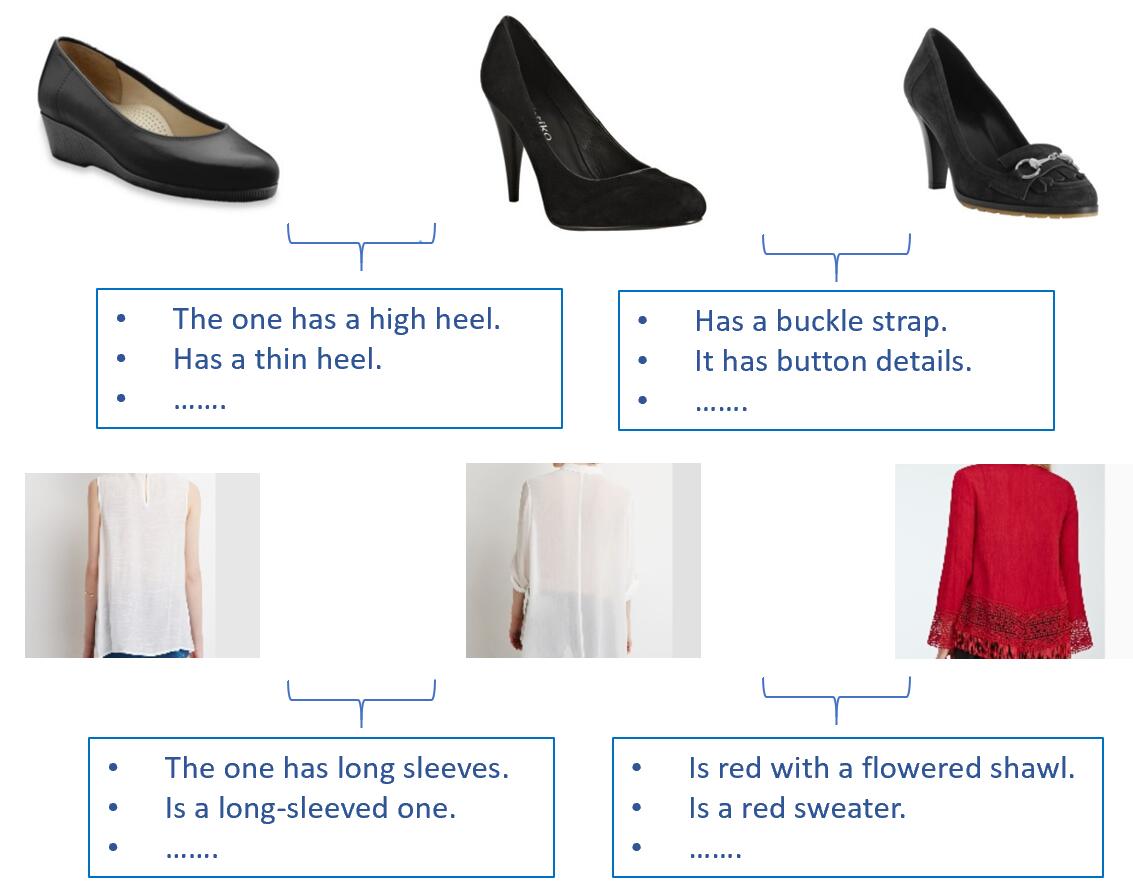}
\caption{\small{Examples of the collected data. Each annotator is asked to provide a natural language sentence describing the difference between two design images. The images and collected descriptions are used to form ``interactive sequences" for the task.} }
\label{fig:data_collection}
\end{figure} 

After manually removing low-quality (e.g., non-descriptive) or duplicate annotations, we obtained a total of 18,497 descriptions for the image sequences from Zap-Seq and 12,765 for DeepFashion-Seq. Table \ref{table:stat:dataset} provides the statistics on the two datasets. Most descriptions are concise (between 4 to 8 words), yet the vocabulary of the description set is diverse (943 unique words in the Zap-Seq dataset and 687 in DeepFashion-Seq). Compared with pre-defined keyword-based attributes provided in the original Zappos and DeepFashion datasets, these natural language descriptions include fine-grained refinement details on the visual design in each image. More details on the datasets (e.g., length distribution of text, phrase-type analysis) can be found in Appendix.


\begin{figure*}[t!]
\centering
\includegraphics[width=0.88\textwidth]{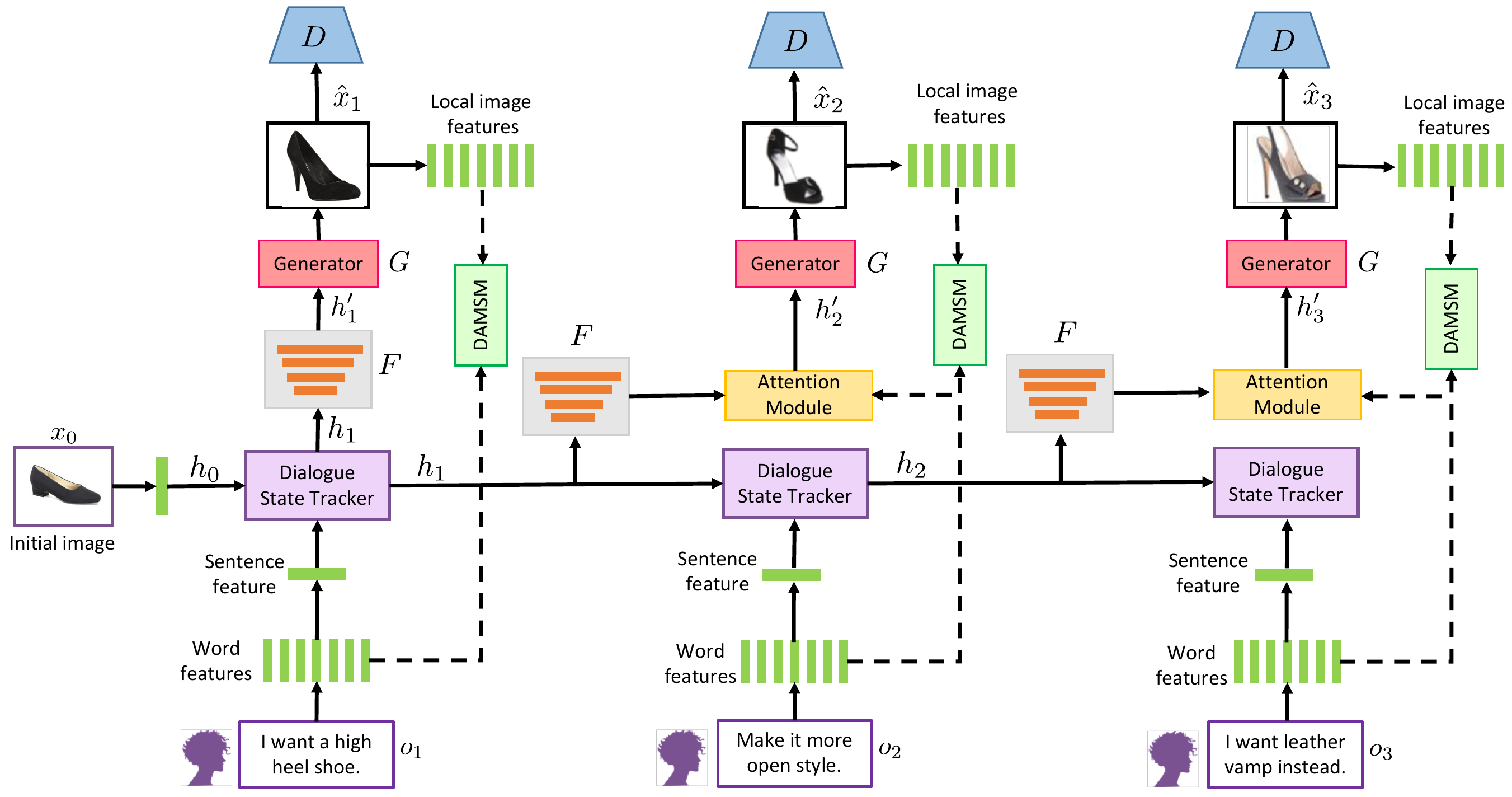}
\caption{\small{The framework of SeqAttnGAN. The neural state tracker keeps track of the contextual information that has been passed on during the sequential image editing process. The attention module absorbs previous context for refining different sub-regions of the image. The DAMSM regularizer provides a fine-grained image-text matching loss. $F$: Up-sampling. $D$: Discriminator.}}
\label{fig:overall_framework}
\end{figure*} 

\section{Sequential Attention GAN} \label{sec:seqattngan}
For this new task, we develop a new Sequential Attention GAN (SeqAttnGAN) model to generate a sequence of images $\hat{x}_1, \ldots, \hat{x}_T$, given an initial input image $x_0$, and a sequence of natural language descriptions $o_1,\ldots, o_T$. The input image $x_0$ is encoded into a feature vector $\mathbf{v}_0$ using ResNet-101 \cite{resnet} pre-trained on ImageNet \cite{imagenet}. Each textual description $o_t$ is encoded via a bi-directional LSTM (BiLSTM), where each word corresponds to two hidden states, one for each direction. We concatenate its two hidden states to obtain the word feature matrix $\mathbf{e}_t \in \R^{d_e\times L}$, where $L$ is the number of words in a sentence, and the $\ell$-th column $\mathbf{e}_t^{(\ell)}$ is the feature vector for the $\ell$-th word. Meanwhile, the last hidden states of the BiLSTM are concatenated into a sentence feature vector, denoted as $\overline{\mathbf{e}}_t \in \R^{d_e}$.

As illustrated in Figure \ref{fig:overall_framework}, in the $t$-th turn ($t\geq 2$), ($i$) the Neural State Tracker fuses the sentence feature vector $\overline{\mathbf{e}}_t$ of the current textual description $o_t$  with the hidden state $\mathbf{h}_{t-1}$, to obtain an updated hidden state $\mathbf{h}_t$; ($ii$) the Attention Module, together with the Up-sampling Module, fuses the word features $\mathbf{e}_t$ of $o_t$ with the feature map that is up-sampled from $\mathbf{h}_{t-1}$, to obtain a context-aware image feature set $\mathbf{h}_t^{\prime}$; ($iii$) the Generator generates the current image $\hat{x}_t$ based on $\mathbf{h}_t^{\prime}$. The following sub-sections introduce each individual component in detail. 

\subsection{Neural State Tracker}
The neural state tracker is modeled as a Recurrent Neural Network (RNN) with the Gated Recurrent Unit (GRU)~\cite{gru}. The initial hidden state $\mathbf{h}_0 = \text{MLP}(\mathbf{v}_0) \in \R^{d_h}$, where $\text{MLP}(\cdot)$ denotes a one-layer MLP layer. In the $t$-th step, the hidden state $\mathbf{h}_t $ is updated via:
\begin{align}
    \mathbf{h}_t &= \mbox{GRU} (\mathbf{h}_{t-1}, \overline{\mathbf{e}}_t)  \,,
\end{align}
where $\mathbf{h}_t$ is considered as the context vector that absorbs all the information from the preceding turns.


\subsection{Attention Module} 
The context vector $\mathbf{h}_{t-1} \in \R^{d_h}$ is first upsampled into a feature map $\mathbf{\tilde{h}}_{t-1} \in \R^{d_h \times N}$, where $N$ is the number of sub-regions in an image. This feature map is then combined with the word feature matrix $\mathbf{e}_t \in \R^{d_e \times L}$ of the current textual description $o_t$ via the Attention Module $F_{\text{attn}}(\cdot)$ to obtain $\mathbf{h}_t^{\prime} \in \R^{d_h \times N}$ , which is used for generating image $\hat{x}_t$. Specifically,
\begin{align}
    \mathbf{h}^{\prime}_{t} = F_{\text{attn}}(\mathbf{e}_t, F(\mathbf{h}_{t-1}))\,, \quad \hat{x}_t = G(\mathbf{h}^{\prime}_{t}, \epsilon_t)\,, 
    \label{eqn:attn_module}
\end{align}
where $F(\cdot)$ is the up-sampling module that transforms  $\mathbf{h}_{t-1}$ into $\mathbf{\tilde{h}}_{t-1}$, $\epsilon_t$ is a noise vector sampled in each step $t$ from a standard normal distribution, and $G(\cdot)$ is the image generator that takes $\mathbf{h_t^{\prime}}$ and $\epsilon_t$ as input. 

The attention module $F_{\text{attn}}$ is used to perform compositional mapping \cite{s2gan,fashiongan,attngan}, i.e., enforcing the model to produce regions and associated features that conform to the textual description.
Specifically, a word-context vector is computed for each sub-region of the image based on its hidden
features $\mathbf{\tilde{h}}_{t-1}$. For the $i$-th sub-region of the image (i.e., the $i$-th column of $\mathbf{\tilde{h}}_{t-1}$, denoted as $\mathbf{\tilde{h}}_{t-1}^{(i)}$), its word-context vector $\mathbf{c}_t^{(i)}$ is obtained via:
\begin{align}
    \bf{c}_t^{(i)} = \sum_{j=0}^{L-1} \beta_{i,j} \bf{e}_t^{(j)}, where\,\,\, \beta_{i,j} = \frac{\exp(s_{i,j})}{\sum_{k=0}^{L-1}\exp(s_{i,k})} \,,
\end{align}
where $s_{i,j} = \mathbf{\tilde{h}}_{t-1}^{(i)} \mathbf{e}_t^{(j)}$, and $\beta_{i,j}$ indicates the weight the model attends to the $j$-th word when generating the $i$-th sub-region of the image.
Finally, the attention module produces a word-context matrix  $\mathbf{h}^{\prime}_{t} = (\mathbf{c}_t^{(0)},\mathbf{c}_t^{(1)},\ldots,\mathbf{c}_t^{(N-1)}) \in \R^{d_{h}\times N}$, which is passed to the image generator $G$ to generate an image $\hat{x}_t$ in the $t$-th step. 

Compared with AttnGAN \cite{attngan}, our model employs the attention module in a sequence, where all the sequence turns share the same image generator $G$ and discriminator $D$, while AttnGAN has disjoint generators and discriminators for different scales. Hence, we name our model \emph{Sequential Attention GAN} (SeqAttnGAN). The objective of SeqAttnGAN is defined as the joint conditional-unconditional losses over the discriminator and the generator. With the supervision of the real image $x_t$ in the $t$-th turn, the loss of the generator $G$ is defined as:
{\small
\begin{equation}
\label{eq:generator}
\mathcal{L}_{G} = -\frac{1}{2} \mathbb{E}_{\hat{x}_{t} \sim P_{G}} [\mathop{\rm{log}}D(\hat{x}_{t})] - \frac{1}{2} \mathbb{E}_{\hat{x}_{t} \sim P_{G}} [\mathop{\rm{log}}D(\hat{x}_{t}, \overline{\mathbf{e}}_t)]\,,
\end{equation}}
and the loss of the discriminator $D$ is calculated by: 
{\small
\begin{equation}
\begin{split}
\label{eq:discriminator}
\mathcal{L}_{D} = -\frac{1}{2} \mathbb{E}_{x_{t} \sim P_{d}} [\mathop{\rm{log}}D(x_t)] - \frac{1}{2} \mathbb{E}_{\hat{x}_{t} \sim P_{G}} [\mathop{\rm{log}}(1-D(\hat{x}_{t}))] \\
-\frac{1}{2} \mathbb{E}_{x_{t} \sim P_{d}} [\mathop{\rm{log}}D(x_{t},\overline{\mathbf{e}}_t)] - \frac{1}{2} \mathbb{E}_{\hat{x}_{t} \sim P_{G}} [\mathop{\rm{log}}(1-D(\hat{x}_t,\overline{\mathbf{e}}_t)]\,,
\end{split}
\end{equation}}
where $x_t$ is from the true data distribution $P_{d}$ and $\hat{x}_{t}$ is from the model distribution $P_{G}$. The above loss is summed over all the sequence turns and paired data samples. 

\subsection{Deep Multimodal Similarity Regularizer}
In addition to the above GAN loss, an image-text matching loss is also introduced into SeqAttnGAN. Specifically, we adopt the Deep Attentional Multimodal Similarity Model (DAMSM) developed in Xu \textit{et al.} \cite{attngan}, which aims to match the similarity between the synthesized images and user input sentences, acting as an effective regularizer
to stabilize the training of the image generator and boost model performance. 

Given a training sample, which is a sequence of $\{x_0, x_1,o_1,\ldots,x_T,o_T\}$, we first transform it into $T$ image-text pairs. Specifically, for each $t=1,\ldots,T$, we use $x_t$ as the input image, and then use the concatenation of the image-attribute value of $x_{t-1}$ (provided in the original datasets) and its associated textual description $o_t$ as the paired text $\hat{o}_t$. Note that here we combine image attributes and user textual input as the new text, which is different from \cite{attngan,stackgan}. In this way, one training sample is transformed into $T$ image-text pairs $\{x_t,\hat{o}_t\}^{T}_{t=1}$. 
Following \cite {dlispr,attngan}, during training, given a mini-batch of $M$ image-text pairs $\{x_i,\hat{o}_i\}^{M}_{i=1}$, the posterior probability of text $\hat{o}_i$ matching image $x_i$ is defined as:
\begin{equation}
\label{eq:dsm:di}
P(\hat{o}_i|x_i) = \frac{ \mathop{\exp} (\gamma R(x_i,\hat{o}_i))}{\sum_{j=1}^{M} \mathop{\exp} (\gamma R(x_i,\hat{o}_j))}\,,
\end{equation}
where $\gamma$ is a smoothing factor, $R(\cdot,\cdot)$ is the word-level attention-driven image-text matching score (i.e., the attention weights are calculated between the sub-region of an image and each word of its corresponding text. See \cite {attngan} for details). The loss function for matching the images with their corresponding text is:
\begin{equation}
\mathcal{L}^{i \to t}_{\text{DAMSM}} = -\textstyle{\sum}^{M}_{i=1}\mathop{\rm{log}}P(\hat{o}_i|x_i)\,.
\end{equation}
Symmetrically, we can also define the loss function for matching textual descriptions with their corresponding images (by switching $\hat{o}_i$ and $x_i$). Combining these two, the regularizer is:
\begin{equation}
\mathcal{L}_{\text{DAMSM}}  = \mathcal{L}^{i \to t}_{\text{DAMSM}} + \mathcal{L}^{t \to i}_{\text{DAMSM}}\,.
\end{equation}
%
By bringing in the discriminative power of the regularizer, the model can generate region-specific image features that better align with the user's text input, as well as improving the visual diversity of generated images.
The final objective of the generator $G$ is defined as:
\begin{equation}
\label{eq:overall}
\mathcal{L} =  \mathcal{L}_{G} + \lambda \mathcal{L}_{\text{DAMSM}}\,,
\end{equation}
where $\lambda$ is the hyperparameter to balance the two loss functions. $\mathcal{L}_{G}$ and $\mathcal{L}_{\text{DAMSM}}$ are summed over all the sequence turns and data samples.

\begin{figure*}[t!]
\centering
\includegraphics[width=0.9\textwidth,height=0.43\textwidth]{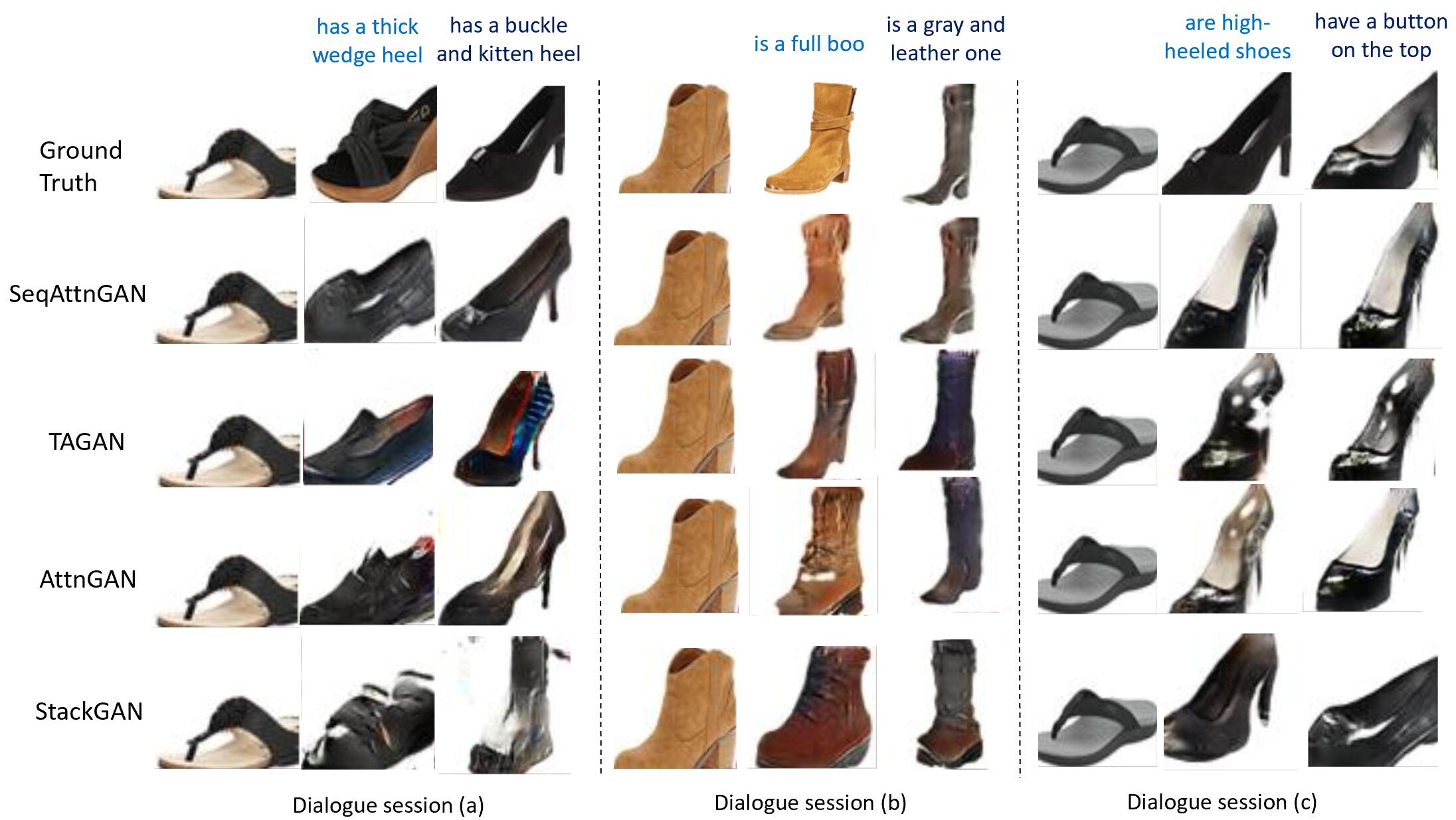}
\caption{\small{Examples of images generated from the given descriptions in the Zap-Seq dataset. The first row shows the ground-truth images and its reference descriptions, followed by images generated by the four approaches: SeqAttnGAN, TAGAN, AttnGAN and StackGAN. To save space, we only display key phrases of each description.}}
\label{fig:shoes_example}
\end{figure*} 

\section{Experiments}
We conduct both quantitative and qualitative evaluations to validate the effectiveness of our proposed model. Given the subjective nature of this new task, human evaluation is also included.
\subsection{Datasets and Baselines}
All the experiments are performed on the Zap-Seq and DeepFashion-Seq datasets with the same splits: 85\% images are used for training, 5\% for validation, and the model is evaluated on a held-out test set from the rest 10\%. We compare our approach with several baselines:
($i$) \textbf{StackGAN} \cite{stackgan} (StackGAN v1 was used due to the relatively low resolution of images in Zap-Seq and DeepFashion-Seq); ($ii$) \textbf{AttnGAN} \cite{attngan}; and  ($iii$) \textbf{TAGAN} \cite{tagan}.
For the three baselines, the hyper-parameter settings and training details remain the same as in the original paper. 

For training, 
data augmentation is used on both datasets. Specifically, images are cropped to $64\times64$ and augmented with horizontal flips. For fair comparison, all models share the same structure of generator and discriminator. Text encoder is also shared. We use the Adam optimizer \cite{KingmaB14} for training. The mini-batch size $M$ is set to 50. $\lambda$ in Eqn. (\ref{eq:overall}) is set to 2 on Zap-Seq and 2.5 on DeepFashion-Seq, respectively. $d_e$ and $d_h$ are set to 300 and 128, respectively. The setting of $\gamma$ follows \cite{attngan}.
DAMSM is used only during training. Baseline model training follows standard conditional-GAN training procedure. 


\subsection{Quantitative Evaluation}
In this section, we provide quantitative evaluation and analysis. For each sequence turn in the test set, we randomly sampled one image from each model, then calculated Inception Score (IS) and Frechet Inception Distance (FID) scores by comparing each selected sample with the ground-truth image. The averaged numbers are presented in Table \ref{table:compared_scores}. On Zap-Seq, SeqAttnGAN performs slightly worse than TAGAN, while on DeepFashion-Seq, our model achieves the best performances. 

Next, to evaluate whether the generated images are coherent with the input text, we measure the Structural Similarity Index (SSIM) score between generated images and ground-truth images. Table \ref{table:ssim} summarizes the results, which show that the images yielded by our model are more consistent with the ground-truth than all the baselines. This indicates that our proposed model can generate images with higher contextual coherency. 

\begin{table}[t!]
\begin{center}
\small
\begin{tabular}{|c|c|c|c|c|}
\hline
Model & \multicolumn{2}{c|}{Zap-Seq} & \multicolumn{2}{c|}{DeepFashion-Seq}\\
\hline 
 & IS & FID & IS & FID \\
\hline \hline
StackGAN & 7.88 & 60.62 & 6.24 & 65.62\\
\hline
AttnGAN & 9.79 & 48.58 & 8.28 & 55.76 \\
\hline
TAGAN & \textbf{9.83} & \textbf{47.25} & 8.26 & 56.49 \\
\hline
SeqAttnGAN  & 9.58 & 50.31 & \textbf{8.41} & \textbf{53.18} \\
\hline
\end{tabular}
\end{center}
\caption{\small{Comparison of Inception Score (IS) and Frechet Inception Distance (FID) between our model and the baselines on the two datasets. IS: higher is better; FID: lower is better.}}\label{table:compared_scores}
\end{table}

\begin{table}
\begin{center}
\small
\begin{adjustbox}{scale=0.9,tabular=|c|c|c|c|c|,center}
\hline 
Dataset & StackGAN & AttnGAN & TAGAN & SeqAttnGAN \\
\hline \hline
Zap-Seq & 0.437 & 0.527 & 0.512 & \textbf{0.651}\\
\hline
DF-Seq & 0.316 & 0.405 & 0.428 & \textbf{0.498}\\
\hline
\end{adjustbox}
\end{center}
\caption{\small{Comparison of SSIM score between our model and the baselines. DF-Seq is short for DeepFashion-Seq.}}\label{table:ssim}
\end{table}

\begin{figure*}
\centering
\includegraphics[width=0.94\textwidth,height=0.43\textwidth]{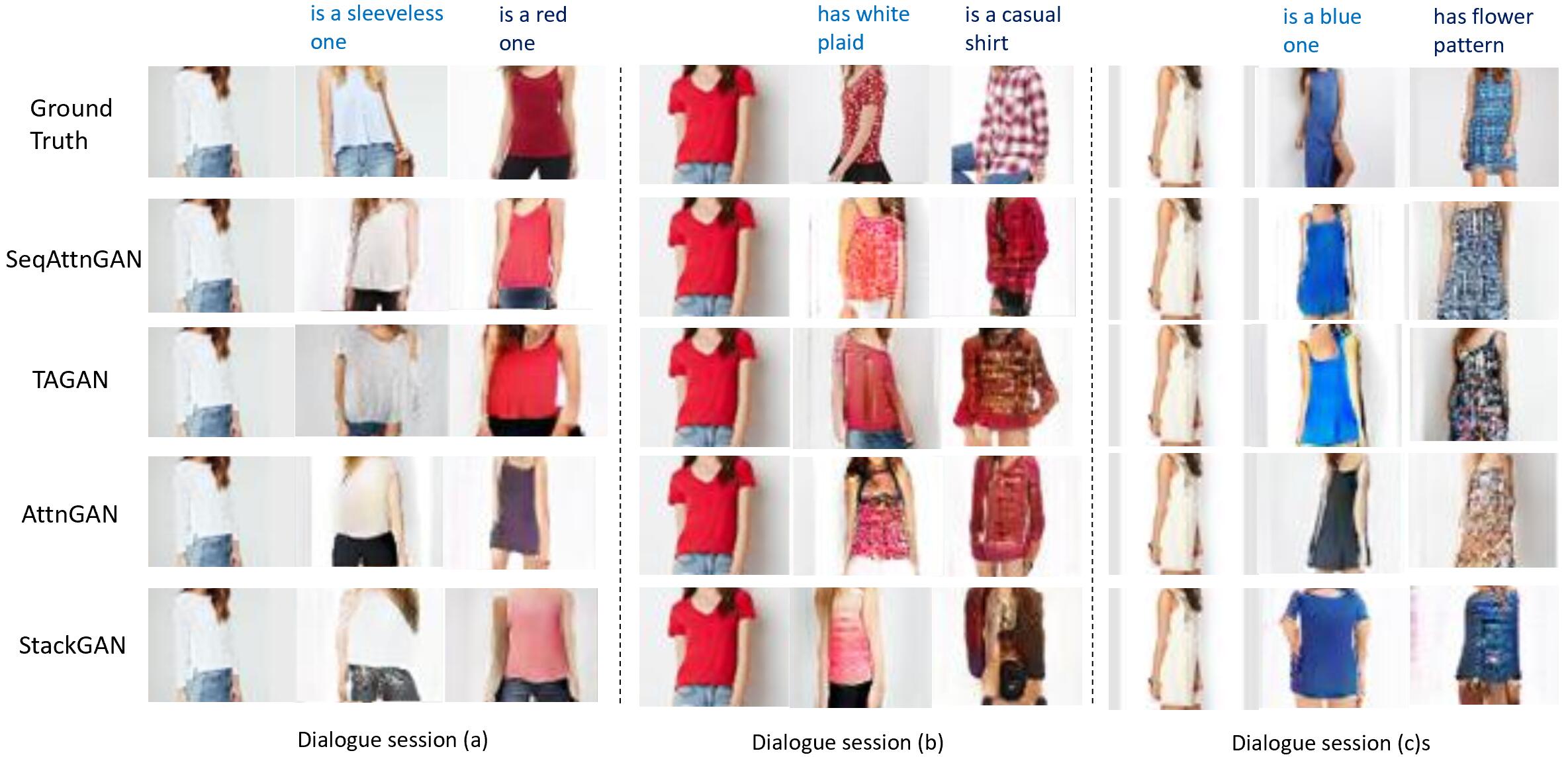}
\caption{\small{Examples of images generated by different methods on the DeepFashions-Seq dataset.}}
\label{fig:clothes_example}
\end{figure*}

Figure \ref{fig:shoes_example} and Figure \ref{fig:clothes_example} present a few examples comparing all the models with the ground-truth (more examples are provided in Appendix). In each example, it is observable that our model generates images more consistent with the ground-truth images as well as the reference descriptions than the baselines. Specifically, SeqAttnGAN can generate: ($i$) better regional changes (e.g., session (a) in Figure \ref{fig:shoes_example}, session (c) in Figure \ref{fig:clothes_example}); and ($ii$) more consistent global changes in color, texture, etc. (e.g., session (b) in Figure \ref{fig:shoes_example}, session (a) in Figure \ref{fig:clothes_example}). Even for fine-grained features (e.g., ``kitten heel'', ``leather'', ``button''), the images generated by our model can well satisfy the requirement. Both AttnGAN and TAGAN are able to synthesize visually sharp/diverse images, but not as good as our model in terms of context relevance (i.e., the generated image does not match the textual description). StackGAN does not perform as well as our model and AttnGAN on either visual quality or image sequence consistency (i.e., the generated image has drastic design change from the previous image). This observation is consistent with the quantitative study. 


\subsection{Human Evaluation}
We also conduct a human evaluation comparing our model with baselines via Amazon Mechanical Turk. From each dataset, we randomly sampled 200 image sequences generated by all the models, each assigned to 3 workers to label. The model from which each image is generated from is hidden from the annotators. The participants were asked to rank the quality of the generated image sequences based on two aspects independently: ($i$) consistency to the given description and the previous image, and ($ii$) visual quality and naturalness.

\begin{table}
\begin{center}
\small
\begin{tabular}{|c|cc|cc|}
\hline
\multirow{2}{*}{Model} & \multicolumn{2}{c|}{Zap-Seq} & \multicolumn{2}{c|}{DeepFashion-Seq} \\
\cline{2-5}
& Vis. & Rel. & Vis. & Rel. \\
\hline \hline
StackGAN & 3.34 & 3.26 & 3.24 & 3.19 \\
\hline
AttnGAN & 2.69 & 2.54 & 2.75 & 2.62 \\
\hline
TAGAN & 2.14 & 2.48 & 2.43 & 2.52  \\
\hline
SeqAttnGAN & \textbf{1.97} & \textbf{1.83} & \textbf{1.70} & \textbf{1.69} \\
\hline
\end{tabular}
\end{center}
\caption{\small{Results from the user study in terms of both visual quality (Vis.) and context relevance (Rel.).
A lower number indicates a higher rank.}}\label{tab:rank}
\end{table}

Table \ref{tab:rank} provides the ranking comparison between SeqAttnGAN and the other methods. For each approach, we computed the average ranking (1 is the best and 4 is the worst). The standard deviation is small, and omitted due to space limit. 
Results show that our approach achieves the best rank on all dimensions, indicating our proposed method achieves the best visual quality and image-text consistency among all the models. 

\begin{figure}
\centering
\includegraphics[width=0.48\textwidth,height=0.3\textwidth]{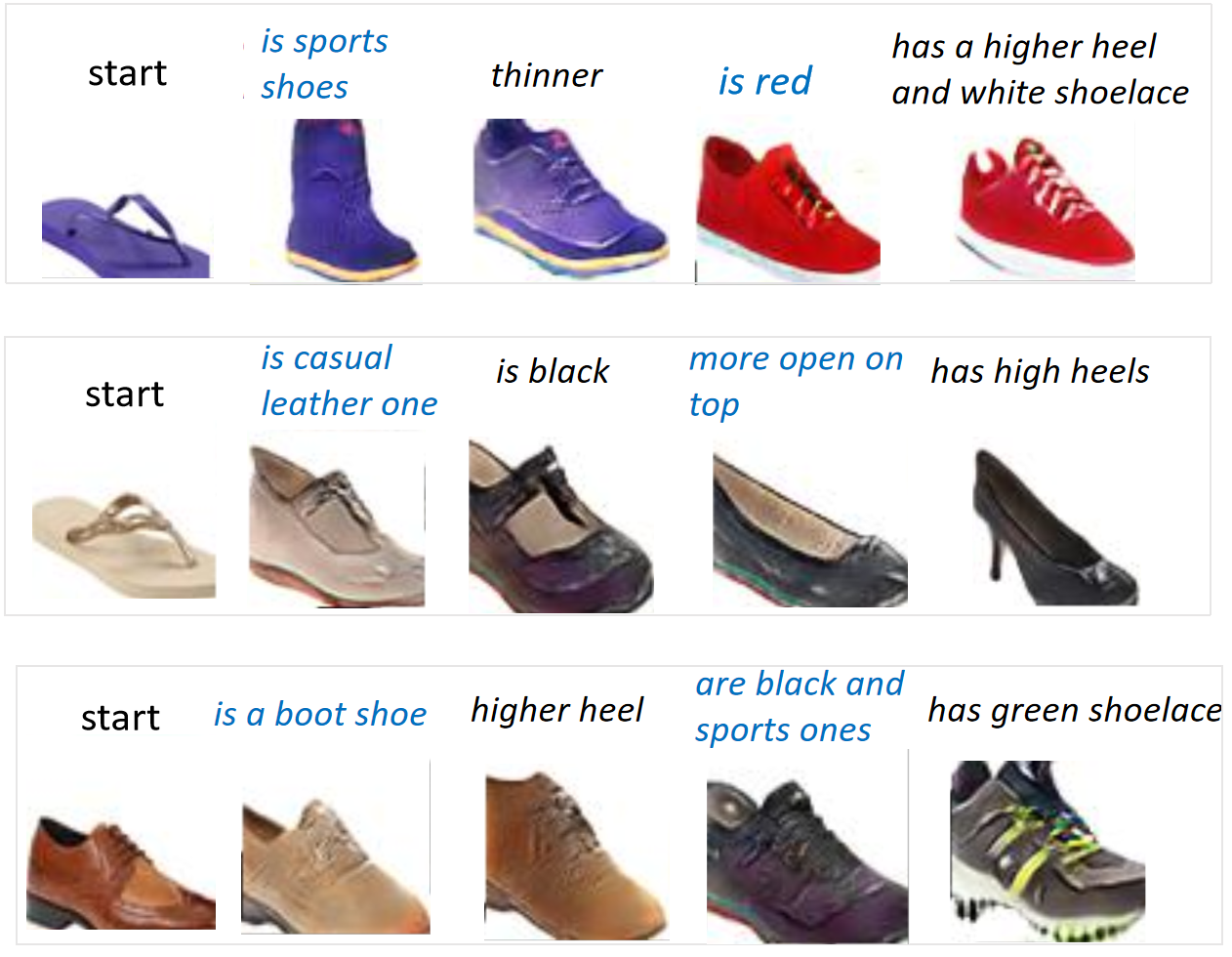}
\caption{Example sessions of users interacting with our image editing system using SeqAttnGAN model. Each row represents an interactive dialogue between the user and our system.}
\label{fig:demo_example}
\end{figure}

Besides the crowdsoured human evaluation, we also recruited real users to interact with our system. Figure \ref{fig:demo_example} shows examples of several dialogue sessions with real users. We observe that users often start the dialogue with a high-level description of main attributes (e,g., color, category). As the dialogue progresses, users gradually give more specific feedback on fine-grained changes. Our model is able to capture both global and subtle changes between images through multi-turn refinement, and can generate high-quality images with fine-grained attributes (e.g., ``white shoelace") as well as comparative descriptions (e.g., ``thinner",``more open"). Overall, these results show potential of applying the proposed SeqAttnGAN model to real-world applications.

\begin{table}[t!]
\begin{center}
\small
\begin{adjustbox}{scale=0.88,tabular=|c|c|c|c|c|c|c|,center}
\hline
Model & \multicolumn{3}{c|}{Zap-Seq} & \multicolumn{3}{c|}{DeepFashion-Seq} \\
\hline \hline
& IS & FID & SSIM & IS & FID & SSIM \\
\hline
SeqAttnGAN & \textbf{9.58} & \textbf{50.31} & \textbf{0.651} & \textbf{8.41} & \textbf{53.18} & \textbf{0.498} \\
\hline
w/o Attn & 8.52 & 57.19 & 0.548 & 7.58 & 58.15 & 0.433 \\
\hline
w/o DAMSM & 8.21 & 58.07 & 0.478 & 7.24 & 60.22 & 0.412 \\
\hline
\end{adjustbox}
\end{center}
\caption{\small{Ablation study on using different variations of SeqAttnGAN, measured by IS, FID and SSIM.} }\label{table:ablation}
\end{table}

\begin{table}
\begin{center}
\small
\begin{adjustbox}{scale=0.9,tabular=|c|c|c|c|,center}
\hline 
 & Single phrase & Composition of phrases & Propositional phrases \\
\hline \hline
IS & 10.13 &  9.82 & 9.41 \\
\hline
FID & 46.59 & 51.02 & 52.78 \\
\hline
\end{adjustbox}
\end{center}
\caption{\small{Ablation study on using different types of phrase on Zap-Seq, measured by IS and FID.}}\label{table:analysis}
\end{table}

\begin{figure}[t!]
\centering
\includegraphics[width=0.45\textwidth]{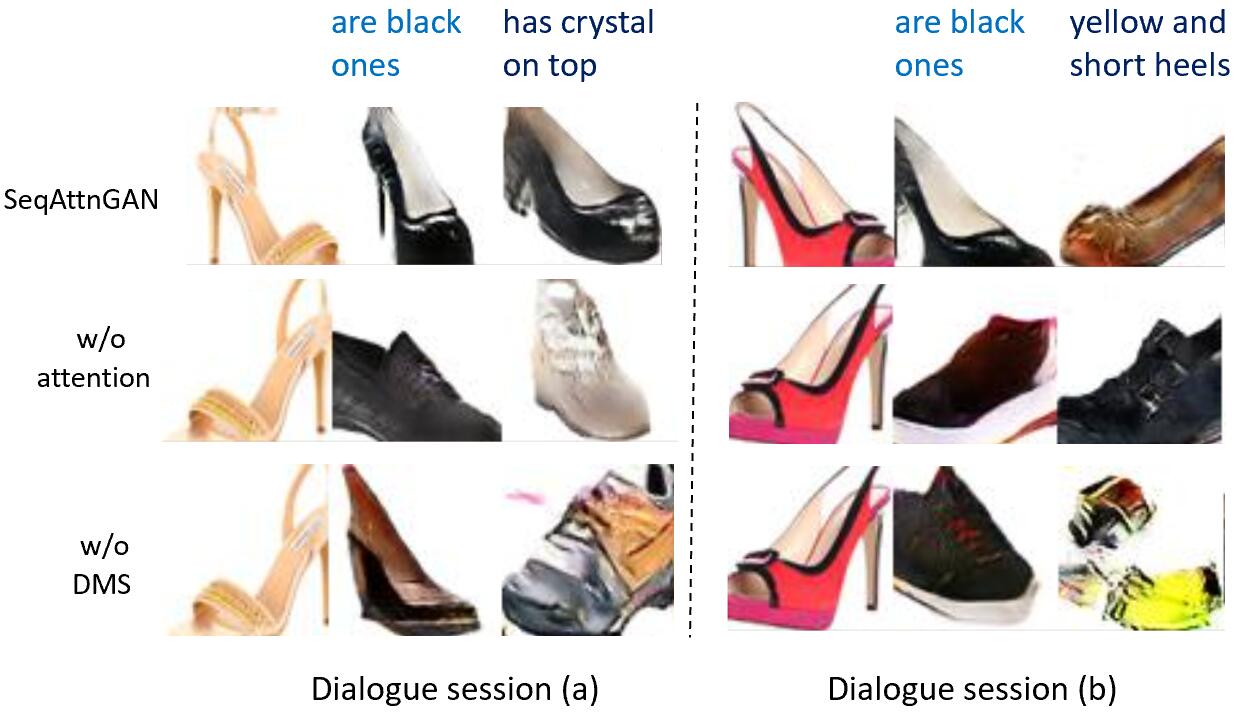}
\caption{\small{Examples generated by different variations of our model. The first row is from SeqAttnGAN, the second row is from SeqAttnGAN without attention, and the last row is from SeqAttnGAN without DAMSM.}}
\label{fig:ablation_example}
\end{figure}

\begin{figure}[t!]
\centering
\includegraphics[width=0.33\textwidth,height=0.12\textwidth]{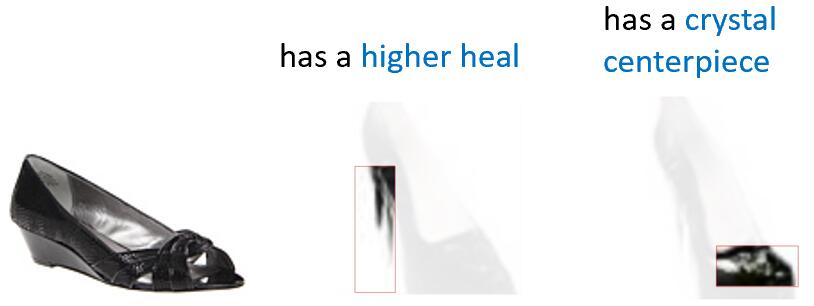}
\caption{\small{An example on the visualization of attended image regions in each step.}}
\label{fig:attention}
\end{figure}

\subsection{Ablation Study}
We conduct ablation study to validate the effectiveness of the two main components in the proposed SeqAttnGAN model: the attention module and the DAMSM regularizer. We first compare the IS, FID and SSIM scores of SeqAttnGAN with/without attention and DAMSM. Table \ref{table:ablation} shows that both attention and DAMSM can improve the model performance with a margin. Figure \ref{fig:ablation_example} provides some examples generated by SeqAttnGAN without attention and DAMSM. As shown in these examples, the ablated models generate images that are drastically different from the previous image, losing contextual consistency, and the textual descriptions are not well reflected in the generated images either.

Figure \ref{fig:attention} provides an illustration of the sequential attention process, where in each step, the targeted region corresponding to the attribute change is attended. This demonstrates that the attention module can help improve image-text consistency. Similar observations can been seen on DeepFashion-Seq as well.

\begin{table*}
\begin{center}
\small
  \begin{tabular}{p{3.3cm} | p{4.5cm} | p{4.8cm}}
    \hline
    {\bf Single Phrase } & \bf{Composition of Phrases } & \bf{Propositional Phrases} \\ 
    {\bf (43\%)} & \bf{(56\%)} & \bf{(35\%)} \\\hline\hline
    
 is brown & is has a buckle and is darker brown & has a strap that goes around the ankle \\ \hline
    has a flatter sole & is closed in and leather & has two more straps across the foot, and slingback strap \\\hline
    has a chevron pattern & has laces and animal print & is a wedge with a strap toe	 \\\hline
    has a reddish lining & has a tall wedge heel, a peep toe, no thong, and a heel strap & has a strap that goes over the foot instead of in the middle \\\hline
  \end{tabular}
\end{center}
 \caption{Examples of captions. More than half of the captions contain composite feedback on more than one types of visual feature.}
\label{tab:phrases}
\end{table*}

We also provide additional quantitative analysis on the Zap-Seq dataset, analysing three types of feedback phrases grouped by text structure (single, composition and propositional phrases). Table \ref{table:analysis} with IS and FID scores on each type shows that single phrases are the easiest to handle, and our model also achieves good scores on the composition of phrases as well as propositional phrases. 

\section{Data Collection and Analysis}
In the section, we explain the details on how we collected Zap-Seq and DeepFashion-Seq datasets and provide insights on the dataset properties. we want the collected descriptions to be concise and relevant for retrieval and avoid casual and non-informative phrases. To this end, as shown in Figure \ref{fig:data_collection}, we design the data collection interface to ask annotators to construct the feedback description. The collection procedure can achieve a balance between lexical flexibility and avoiding irrelevant phrases. After manual data cleaning, we are able to get 18,497 descriptions for Zap-Seq and 12,765 for DeepFashion-Seq. 

\begin{figure}
\centering
\includegraphics[width=0.48\textwidth,height=0.23\textwidth]{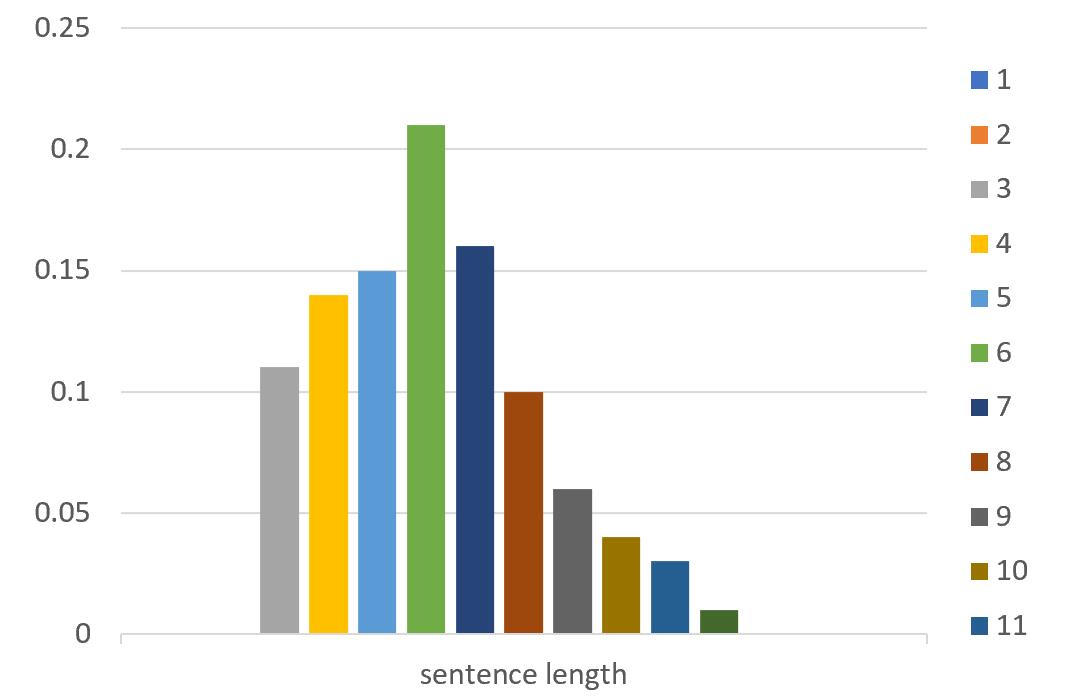}
\caption{Length distribution of the caption data.}
\label{fig:data_distribution}
\end{figure}

Figure \ref{fig:data_distribution} shows the length distribution of the collected captions. Most captions are very concise (3-8 words). In Table \ref{tab:phrases}, we provide some examples in different types. We can see many feedback expressions consist of compositions of multiple phrases, including spatial or structural details. Understanding the semantic meaning of the captions is one of the most challenging part for this task.

\begin{figure}
\centering
\includegraphics[width=0.48\textwidth,height=0.23\textwidth]{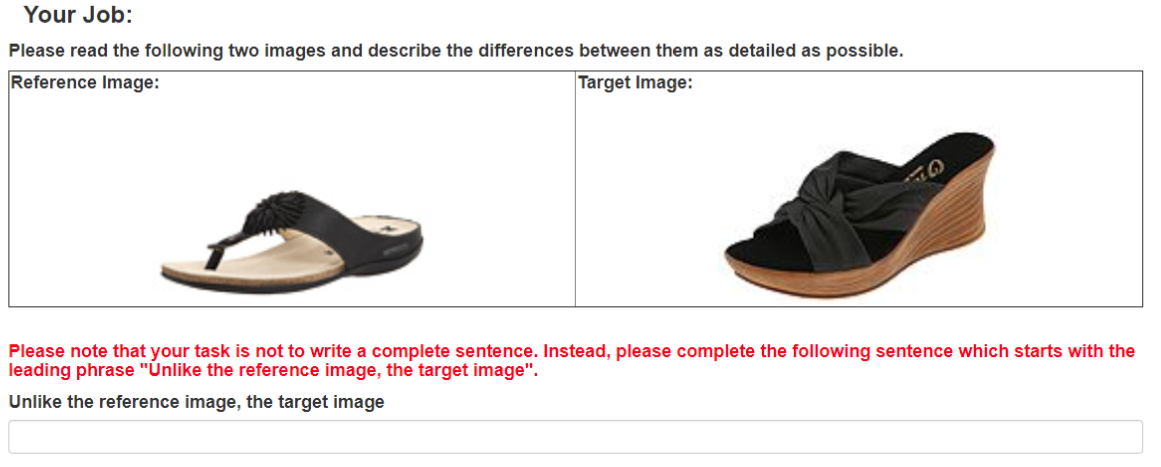}
\caption{AMT annotation interface. Annotators need to complete the rest of the response message given the scenario.}
\label{fig:data_collection}
\end{figure}

\section{Conclusion}
We present interactive image editing, a novel task that resides in the intersection of computer vision and language. To provide benchmarks, we introduce two new datasets, Zap-Seq and DeepFashion-Seq, which contain image sequences accompanied by textual descriptions. A SeqAttnGAN model is proposed for this task. Experiments on the two datasets demonstrate that SeqAttnGAN outperforms baseline methods across visual quality, image-text relevance and image sequence consistency. 
For future work, we plan to apply the proposed model to other image types and explore how to generate more consistent image sequences by disentangling learned representations into attributes and other factors. In addition, understanding semantic meanings (e.g., ``in front", ``on side") of user feedbacks is also important. We also plan to investigate models to support more robust natural language interactions, which requires techniques such as user intent understanding and co-reference resolution. 



\clearpage
\bibliographystyle{abbrv}
{\small
\bibliography{egbib}}

\begin{thebibliography}{10}

\bibitem{VQA}
S.~Antol, A.~Agrawal, J.~Lu, M.~Mitchell, D.~Batra, C.~L. Zitnick, and
  D.~Parikh.
\newblock Vqa: Visual question answering.
\newblock In {\em ICCV}, 2015.

\bibitem{ashual2019specifying}
O.~Ashual and L.~Wolf.
\newblock Specifying object attributes and relations in interactive scene
  generation.
\newblock In {\em ICCV}, 2019.

\bibitem{Benmalek2018TheNP}
R.~Y. Benmalek, C.~Cardie, S.~J. Belongie, X.~He, and J.~Gao.
\newblock The neural painter: Multi-turn image generation.
\newblock {\em arXiv preprint arXiv:1806.06183}, 2018.

\bibitem{Bordes2016LearningEG}
A.~Bordes, Y.-L. Boureau, and J.~Weston.
\newblock Learning end-to-end goal-oriented dialog.
\newblock In {\em ICLR}, 2017.

\bibitem{amt}
M.~Buhrmester, T.~Kwang, and S.~Gosling.
\newblock Amazon's mechanical turk: A new source of inexpensive, yet
  high-quality, data?
\newblock {\em Perspectives on Psychological Science}, 2011.

\bibitem{libe}
J.~Chen, Y.~Shen, J.~Gao, J.~Liu, and X.~Liu.
\newblock Language-based image editing with recurrent attentive models.
\newblock In {\em CVPR}, 2018.

\bibitem{gru}
J.~Chung, C.~Gulcehre, K.~Cho, and Y.~Bengio.
\newblock Empirical evaluation of gated recurrent neural networks on sequence
  modeling.
\newblock In {\em NeurIPS Workshop}, 2014.

\bibitem{das2017visual}
A.~Das, S.~Kottur, K.~Gupta, A.~Singh, D.~Yadav, J.~M. Moura, D.~Parikh, and
  D.~Batra.
\newblock Visual dialog.
\newblock In {\em CVPR}, 2017.

\bibitem{DBLP:conf/cvpr/VriesSCPLC17}
H.~de~Vries, F.~Strub, S.~Chandar, O.~Pietquin, H.~Larochelle, and A.~C.
  Courville.
\newblock Guesswhat?! visual object discovery through multi-modal dialogue.
\newblock In {\em CVPR}, 2017.

\bibitem{imagenet}
J.~Deng, W.~Dong, R.~Socher, L.-J. Li, K.~Li, and L.~Fei-Fei.
\newblock {ImageNet: A Large-Scale Hierarchical Image Database}.
\newblock In {\em CVPR}, 2009.

\bibitem{Dixit_Kwitt_Niethammer_Vasconcelos_2017}
M.~Dixit, R.~Kwitt, M.~Niethammer, and N.~Vasconcelos.
\newblock Aga: Attribute guided augmentation.
\newblock In {\em CVPR}, 2017.

\bibitem{gaosurvey}
J.~Gao, M.~Galley, and L.~Li.
\newblock Neural approaches to conversational ai.
\newblock {\em arXiv preprint arXiv:1809.08267}, 2018.

\bibitem{gan}
I.~Goodfellow, J.~Pouget-Abadie, M.~Mirza, B.~Xu, D.~Warde-Farley, S.~Ozair,
  A.~Courville, and Y.~Bengio.
\newblock Generative adversarial nets.
\newblock In {\em NeurIPS}, 2014.

\bibitem{DBLP:journals/corr/abs-1805-00145}
X.~Guo, H.~Wu, Y.~Cheng, S.~Rennie, and R.~S. Feris.
\newblock Dialog-based interactive image retrieval.
\newblock In {\em NeurIPS}, 2018.

\bibitem{resnet}
K.~He, X.~Zhang, S.~Ren, and J.~Sun.
\newblock Deep residual learning for image recognition.
\newblock In {\em CVPR}, 2016.

\bibitem{dlispr}
X.~He, L.~Deng, and W.~Chou.
\newblock Discriminative learning in sequential pattern recognition.
\newblock {\em IEEE Signal Processing Magazine}, 2008.

\bibitem{hu2016segmentation}
R.~Hu, M.~Rohrbach, and T.~Darrell.
\newblock Segmentation from natural language expressions.
\newblock In {\em ECCV}, 2016.

\bibitem{imagetrans}
P.~Isola, J.-Y. Zhu, T.~Zhou, and A.~A. Efros.
\newblock Image-to-image translation with conditional adversarial networks.
\newblock In {\em CVPR}, 2017.

\bibitem{sg2im}
J.~Johnson, A.~Gupta, and L.~Fei-Fei.
\newblock Image generation from scene graphs.
\newblock In {\em CVPR}, 2018.

\bibitem{codraw}
J.~Kim, D.~Parikh, D.~Batra, B.~Zhang, and Y.~Tian.
\newblock Codraw: Visual dialog for collaborative drawing.
\newblock {\em CoRR}, abs/1712.05558, 2017.

\bibitem{KingmaB14}
D.~P. Kingma and J.~Ba.
\newblock Adam: {A} method for stochastic optimization.
\newblock {\em arXiv preprint arXiv:1412.6980}, 2014.

\bibitem{mmdgan}
C.-L. Li, W.-C. Chang, Y.~Cheng, Y.~Yang, and B.~Poczos.
\newblock Mmd gan: Towards deeper understanding of moment matching network.
\newblock In {\em NeurIPS}, pages 2203--2213, 2017.

\bibitem{li2020bachgan}
Y.~Li, Y.~Cheng, Z.~Gan, L.~Yu, L.~Wang, and J.~Liu.
\newblock Bachgan: High-resolution image synthesis from salient object layout.
\newblock {\em arXiv preprint arXiv:2003.11690}, 2020.

\bibitem{li2018storygan}
Y.~Li, Z.~Gan, Y.~Shen, J.~Liu, Y.~Cheng, Y.~Wu, L.~Carin, D.~Carlson, and
  J.~Gao.
\newblock Storygan: A sequential conditional gan for story visualization.
\newblock In {\em CVPR}, 2019.

\bibitem{Liu2017UIT}
M.-Y. Liu, T.~Breuel, and J.~Kautz.
\newblock Unsupervised image-to-image translation networks.
\newblock In {\em NeurIPS}, 2017.

\bibitem{liuLQWTcvpr16DeepFashion}
Z.~Liu, P.~Luo, S.~Qiu, X.~Wang, and X.~Tang.
\newblock Deepfashion: Powering robust clothes recognition and retrieval with
  rich annotations.
\newblock In {\em CVPR}, 2016.

\bibitem{manuvinakurike_conversational_2018}
R.~Manuvinakurike, T.~Bui, W.~Chang, and K.~Georgila.
\newblock Conversational {Image} {Editing}: {Incremental} {Intent}
  {Identiﬁcation} in a {New} {Dialogue} {Task}.
\newblock In {\em SIGDIAL}, Melbourne, Australia, 2018.

\bibitem{manuvinakurike_using_2017}
R.~Manuvinakurike, D.~DeVault, and K.~Georgila.
\newblock Using {Reinforcement} {Learning} to {Model} {Incrementality} in a
  {Fast}-{Paced} {Dialogue} {Game}.
\newblock In {\em SIGDIAL}, Saarbruecken Germany, 2017.

\bibitem{Mirza2014ConditionalGA}
M.~Mirza and S.~Osindero.
\newblock Conditional generative adversarial nets.
\newblock {\em arXiv preprint arXiv:1411.1784}, 2014.

\bibitem{mostafazadeh2017image}
N.~Mostafazadeh, C.~Brockett, B.~Dolan, M.~Galley, J.~Gao, G.~Spithourakis, and
  L.~Vanderwende.
\newblock Image-grounded conversations: Multimodal context for natural question
  and response generation.
\newblock In {\em IJCNLP}, 2017.

\bibitem{tagan}
S.~Nam, Y.~Kim, and S.~J. Kim.
\newblock Text-adaptive generative adversarial networks: Manipulating images
  with natural language.
\newblock In {\em NeurIPS}, 2018.

\bibitem{pmlr-v70-odena17a}
A.~Odena, C.~Olah, and J.~Shlens.
\newblock Conditional image synthesis with auxiliary classifier {GAN}s.
\newblock In {\em ICML}, 2017.

\bibitem{pathakCVPR16context}
D.~Pathak, P.~Kr\"ahenb\"uhl, J.~Donahue, T.~Darrell, and A.~Efros.
\newblock Context encoders: Feature learning by inpainting.
\newblock In {\em CVPR}, 2016.

\bibitem{pmlr-v48-reed16}
S.~Reed, Z.~Akata, X.~Yan, L.~Logeswaran, B.~Schiele, and H.~Lee.
\newblock Generative adversarial text to image synthesis.
\newblock In {\em ICML}, 2016.

\bibitem{rohrbach16eccv}
A.~Rohrbach, M.~Rohrbach, R.~Hu, T.~Darrell, and B.~Schiele.
\newblock Grounding of textual phrases in images by reconstruction.
\newblock In {\em ECCV}, 2016.

\bibitem{sangkloy2016scribbler}
P.~Sangkloy, J.~Lu, C.~Fang, F.~Yu, and J.~Hays.
\newblock Scribbler: Controlling deep image synthesis with sketch and color.
\newblock In {\em CVPR}, 2017.

\bibitem{Serban:2016:BED:3016387.3016435}
I.~V. Serban, A.~Sordoni, Y.~Bengio, A.~Courville, and J.~Pineau.
\newblock Building end-to-end dialogue systems using generative hierarchical
  neural network models.
\newblock In {\em AAAI}, 2016.

\bibitem{chatpainter}
S.~Sharma, D.~Suhubdy, V.~Michalski, S.~E. Kahou, and Y.~Bengio.
\newblock Chatpainter: Improving text to image generation using dialogue.
\newblock {\em CoRR}, abs/1802.08216, 2018.

\bibitem{wang2016learning}
L.~Wang, Y.~Li, and S.~Lazebnik.
\newblock Learning deep structure-preserving image-text embeddings.
\newblock In {\em CVPR}, 2016.

\bibitem{s2gan}
X.~Wang and A.~Gupta.
\newblock Generative image modeling using style and structure adversarial
  networks.
\newblock In {\em ECCV}, 2016.

\bibitem{xian2017texturegan}
W.~Xian, P.~Sangkloy, V.~Agrawal, A.~Raj, J.~Lu, C.~Fang, F.~Yu, and J.~Hays.
\newblock Texturegan: Controlling deep image synthesis with texture patches.
\newblock {\em arXiv preprint arXiv:1706.02823}, 2017.

\bibitem{attngan}
T.~Xu, P.~Zhang, Q.~Huang, H.~Zhang, Z.~Gan, X.~Huang, and X.~He.
\newblock Attngan: Fine-grained text to image generation with attentional
  generative adversarial networks.
\newblock In {\em CVPR}, 2018.

\bibitem{Yu:2014:FVC:2679600.2680101}
A.~Yu and K.~Grauman.
\newblock Fine-grained visual comparisons with local learning.
\newblock In {\em CVPR}, 2014.

\bibitem{stackgan}
H.~Zhang, T.~Xu, H.~Li, S.~Zhang, X.~Wang, X.~Huang, and D.~Metaxas.
\newblock Stackgan: Text to photo-realistic image synthesis with stacked
  generative adversarial networks.
\newblock In {\em ICCV}, 2017.

\bibitem{Zhao2017MemoryAugmentedAM}
B.~Zhao, J.~Feng, X.~Wu, and S.~Yan.
\newblock Memory-augmented attribute manipulation networks for interactive
  fashion search.
\newblock In {\em CVPR}, 2017.

\bibitem{zhao2019image}
B.~Zhao, L.~Meng, W.~Yin, and L.~Sigal.
\newblock Image generation from layout.
\newblock In {\em CVPR}, 2019.

\bibitem{CycleGAN2017}
J.-Y. Zhu, T.~Park, P.~Isola, and A.~A. Efros.
\newblock Unpaired image-to-image translation using cycle-consistent
  adversarial networks.
\newblock In {\em ICCV}, 2017.

\bibitem{NIPS2017_6650}
J.-Y. Zhu, R.~Zhang, D.~Pathak, T.~Darrell, A.~A. Efros, O.~Wang, and
  E.~Shechtman.
\newblock Toward multimodal image-to-image translation.
\newblock In {\em NeurIPS}, 2017.

\bibitem{fashiongan}
S.~Zhu, S.~Fidler, R.~Urtasun, D.~Lin, and C.~C. Loy.
\newblock Be your own prada: Fashion synthesis with structural coherence.
\newblock In {\em ICCV}, 2017.

\end{thebibliography}
\end{document}